\documentclass[10pt,twocolumn]{article}

\usepackage[margin=0.75in]{geometry}
\usepackage{microtype}

\usepackage{graphicx}
\usepackage{subcaption}
\usepackage{booktabs}
\usepackage{multirow}
\usepackage{siunitx}
\usepackage{makecell}
\usepackage{float}
\usepackage{placeins}
\usepackage[numbers,sort&compress]{natbib}

\usepackage{amsmath}
\usepackage{amssymb}
\usepackage{mathtools}
\usepackage{amsthm}

\usepackage[colorlinks=true,linkcolor=blue,citecolor=blue,urlcolor=blue]{hyperref}
\usepackage[capitalize,noabbrev]{cleveref}

\usepackage[disable]{todonotes}

\theoremstyle{plain}

\theoremstyle{definition}

\theoremstyle{remark}

\title{\bfseries BiTimeCrossNet: Time-Aware Self-Supervised Learning for Pediatric Sleep}
\author{
Saurav Raj Pandey$^{1}$ \and
Harlin Lee$^{2}$ \\[0.5em]
\begin{minipage}{\linewidth}
\centering
\small
$^{1}$Department of Computer Science, UNC Chapel Hill, Chapel Hill, NC, USA \\
$^{2}$School of Data Science and Society, UNC Chapel Hill, Chapel Hill, NC, USA \\[0.3em]
\texttt{srpandey@unc.edu},\ \texttt{harlin@unc.edu}
\end{minipage}
}

\date{}

\begin{document}

\twocolumn[
\maketitle
\vspace{-0.4cm}
\vspace{0.2cm}
]
\begin{abstract}
We present BiTimeCrossNet (BTCNet), a multimodal self-supervised learning framework for long physiological recordings such as overnight sleep studies. While many existing approaches train on short segments treated as independent samples, BTCNet incorporates information about when each segment occurs within its parent recording, for example within a sleep session. BTCNet further learns pairwise interactions between physiological signals via cross-attention, without requiring task labels or sequence-level supervision.

We evaluate BTCNet on pediatric sleep data across six downstream tasks, including sleep staging, arousal detection, and respiratory event detection. Under frozen-backbone linear probing, BTCNet consistently outperforms an otherwise identical non–time-aware variant, with gains that generalize to an independent pediatric dataset. Compared to existing multimodal self-supervised sleep models, BTCNet achieves strong performance, particularly on respiration-related tasks.
\end{abstract}

\section{Introduction}
Sleep plays a critical role in physical and mental health in childhood. Disrupted sleep has been associated with adverse outcomes including anxiety, depression, hypertension, impaired academic performance, and behavioral difficulties \cite{sleep_disruption}. In clinical practice, sleep is most commonly assessed using polysomnography (PSG), an overnight study that records multiple physiological signals such as electroencephalography (EEG), electrooculography (EOG), and respiratory and cardiac measurements. These recordings are manually reviewed and annotated by trained sleep technicians to identify sleep stages and clinically relevant events, including oxygen desaturations and respiratory disturbances. While effective, this process is labor-intensive, costly, and difficult to scale.

\begin{figure}[t]
    \centering
    \includegraphics[width=\linewidth]{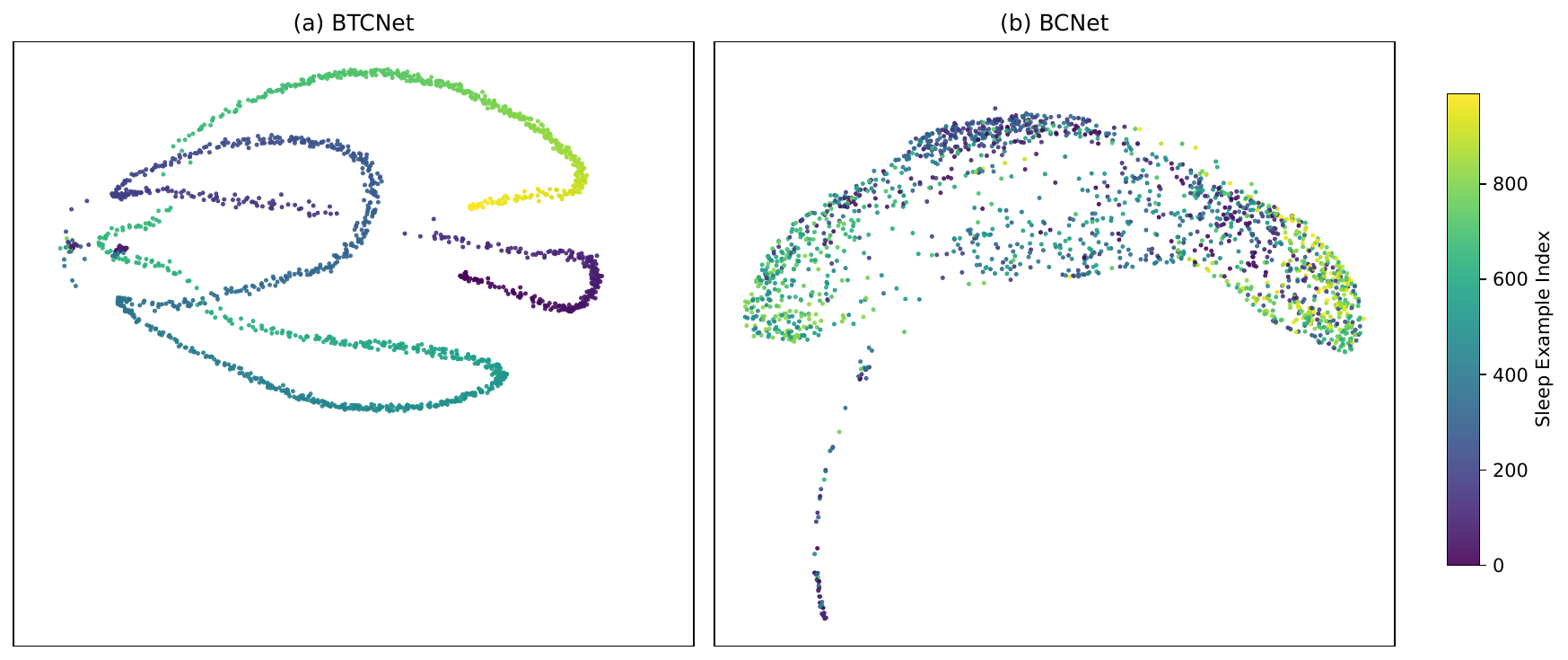}
    \caption{
    PHATE visualization \cite{moon2019visualizing} of frozen embeddings from BTCNet (time-aware) and BCNet (non-time-aware) for a representative patient using SPO2 and CAPNO signals. Each point is 30 seconds of sleep colored by time of night. The time-aware model exhibits a smoother and more coherent global structure compared to the non-time-aware model.
    }
    \label{fig:phate}
\end{figure}
Recent advances in machine learning offer the potential to automate large portions of PSG analysis. However, models trained primarily on adult sleep data generalize poorly to pediatric populations, even for fundamental tasks such as sleep stage classification \cite{nazih2023influence}. This performance gap reflects well-documented physiological differences between pediatric and adult sleep \cite{accardo2010differences, owens2012pro}. As a result, pediatric sleep analysis requires specific data and models that can capture age-specific sleep dynamics.

Self-supervised learning (SSL) provides a promising framework for learning representations from unlabeled PSG recordings using pretext objectives such as masked signal reconstruction. These representations can then be transferred to downstream tasks with substantially reduced annotation requirements. While SSL has achieved notable success in adult sleep modeling, its adoption in pediatric sleep remains limited. Moreover, existing SSL approaches are largely optimized for sleep stage classification \cite{UNet, wenjian2023dynamicsleepnet, lee2024neuronet, kostas2021bendr}. In contrast, clinically important diagnostic tasks, such as apnea, hypopnea, and oxygen desaturation detection, have received comparatively little attention, despite their strong associations with cardiovascular and neurological outcomes \cite{yaggi2010adult}.

A further limitation of existing SSL formulations is their reliance on short, fixed-duration temporal windows, for example 30-second epochs. Although convenient, this design ignores the fact that clinically meaningful sleep physiology evolves over the course of an entire night (Figure~\ref{fig:phate}). Respiratory stability, autonomic tone, and heart rate dynamics vary systematically across sleep cycles and circadian time \cite{choudhary2009sleep, stein2012heart}. Nevertheless, most SSL methods treat sleep windows as temporally independent samples, without encoding information about where a given window occurs within the broader trajectory of a sleep session.

\paragraph{Main Contributions}
We introduce \textbf{BiTimeCrossNet (BTCNet)}, a multimodal self-supervised learning (SSL) framework for pediatric polysomnography (PSG) that learns physiologically meaningful representations from heterogeneous signals with minimal manual annotation. BTCNet is pretrained on the Nationwide Children's Hospital (NCH) Sleep Databank \cite{lee2022large} using a hybrid SSL objective that combines masked autoencoder \cite{he2022masked} with contrastive learning. Our contributions are as follows:

\begin{itemize}
    \item We propose \textbf{BTCNet}, a pediatric multimodal SSL framework that learns transferable representations across heterogeneous physiological channels and supports \textbf{six clinically relevant downstream tasks}.

    \item We introduce a novel \textbf{random modality-pair cross-attention pretraining strategy}, in which pairs (hence the prefix ``Bi'') of physiological signals are randomly sampled at each iteration. This strategy encourages robust learning of inter-modal dependencies through a cross-attention mechanism and improves resilience to missing or noisy modalities.

\item   We present a \textbf{global time-aware positional conditioning mechanism} that injects night-scale temporal context into transformer representations, enabling modeling of physiological dynamics over an entire sleep session. To our knowledge, this is the first incorporation of global time-of-sleep context into SSL for sleep analysis. We show that \textbf{time-aware pretraining consistently improves} AUROC and F1 across \textbf{all six tasks} and multiple modality combinations on NCH and an independent pediatric cohort (CHAT).

  \item We demonstrate \textbf{cross-dataset generalization} by evaluating BTCNet out of the box on CHAT via linear probing, where it outperforms recent publicly available multimodal SSL models on oxygen desaturation and apnea-related detection tasks.

\end{itemize}

\section{Related Work}
\label{sec:related_work}
Self-supervised learning (SSL) has become a dominant paradigm for learning transferable representations across domains. In the sleep domain, SSL efforts have mostly focused on uni-modal EEG-based representation learning, with a strong emphasis on sleep-stage classification, including BENDR \cite{kostas2021bendr}, EEGPT \cite{wang2024eegpt}, and NeuroNet \cite{lee2024neuronet}. Recent work has also explored multimodal SSL for sleep analysis, such as COCOA \cite{deldari2022cocoa}, SleepFM \cite{sleepfm_original}, and SynthSleepNet \cite{CASleepNet}. However, multimodal SSL techniques trained to analyze pediatric sleep are further limited. To our knowledge, PedSleepMAE \cite{pandey2024pedsleepmae} is the only existing SSL framework trained on large-scale pediatric PSG data (NCH Sleep Databank) for sleep-stage classification and related diagnostic tasks (e.g., apnea and hypopnea detection).

Recent work by \cite{ye2025uncovering} demonstrates that session-level trajectory and topological features extracted from frozen PedSleepMAE embeddings can improve diagnostic performance. These findings suggest that global time-of-sleep structure contains clinically meaningful information, motivating our approach to incorporate time-of-sleep context directly into self-supervised pretraining rather than as a downstream augmentation.

BTCNet differs from most prior self-supervised sleep representation learning methods in several key ways.
First, MAE-based approaches such as PedSleepMAE, NeuroNet, and SynthSleepNet rely on standard local positional encodings and do not explicitly model global sleep-level temporal context during pretraining. In contrast, BTCNet, in addition to positional encodings, conditions representations on where a sleep window occurs within its full sleep session. Second, while most prior multimodal SSL methods are trained and evaluated under predefined modality configurations, BTCNet uses randomly sampled modality pairs with shared cross-attention, enabling flexible and task-specific channel selection (two channels only) at evaluation time. Finally, unlike SleepFM, which emphasizes missing-modality robustness via global modality alignment, BTCNet focuses on learning pairwise physiological interactions through cross-attention.

\section{Methods}

\begin{figure}[htbp]
    \centering
\includegraphics[
  width=1.05\linewidth,
  trim=0.5cm 0cm 0.5cm 0cm,
  clip
]{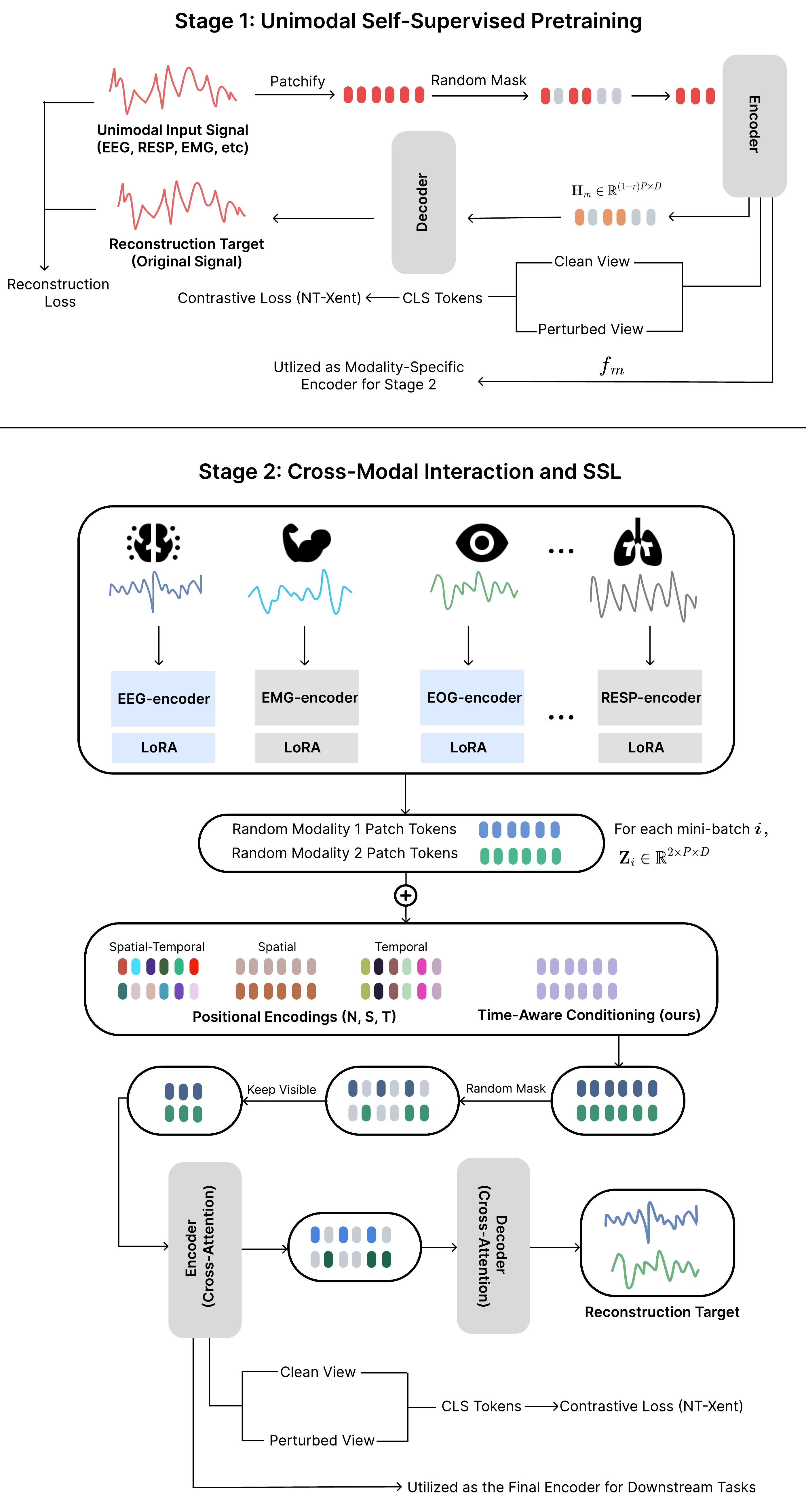}
    \caption{
    Overview of BTCNet. 
    \textbf{Stage 1:} Unimodal self-supervised pretraining using masked reconstruction and contrastive learning to obtain modality-specific encoders.
    \textbf{Stage 2:} Cross-modal self-supervised learning with randomly sampled modality pairs, time-aware conditioning, and cross-attention.
    }
    \label{fig:btcnet}
\end{figure}

We describe the dataset and the architecture of BTCNet along with our novel time-aware mechanism. BTCNet will be made open source upon acceptance of the paper. 

\subsection{Dataset and Preprocessing}
\subsubsection{NCH Sleep Databank}
BTCNet is pretrained on the Nationwide Children’s Hospital (NCH) Sleep Databank, a large publicly available pediatric polysomnography (PSG) dataset collected in real-world clinical settings \cite{lee2022large}. We analyze 2{,}379 overnight PSG recordings containing a consistent set of sixteen commonly available physiological channels, including EEG, EOG, and respiratory signals.

Expert annotations are provided at 30-second resolution, including sleep stages and clinically relevant events such as apnea, hypopnea, arousal, and oxygen desaturation. Each 30-second epoch is treated as an individual training example. Additional details are in Appendix~\ref{app:nch_dataset}.

\subsubsection{CHAT Dataset}
The Childhood Adenotonsillectomy Trial (CHAT) is a multi-center randomized controlled study of children with mild to moderate obstructive sleep apnea, with standardized overnight polysomnography and centralized scoring \cite{zhang2018national, marcus2013randomized}. We use the baseline subset collected prior to any intervention. 

To ensure compatibility with NCH, we restrict CHAT to the twelve overlapping physiological channels, which comprises of 422 PSGs. The CHAT dataset is used exclusively for external validation and cross-dataset generalization and is not used during BTCNet pretraining. Additional details are provided in Appendix~\ref{app:chat_dataset}.

\subsection{BTCNet Architecture}

In this section, we discuss the architecture of model \textbf{BTCNet}. BTCNet is trained in two stages: (1) unimodal self-supervised pretraining for each individual physiological modality, and (2) bimodal cross-modal pretraining using randomly sampled modality pairs. The second stage augments the signals with our novel time-aware structure and cross-attention, followed by self-supervised learning, enabling BTCNet to capture both modality-specific features and inter-modal dependencies. Figure \ref{fig:btcnet} displays this setup in detail.

\subsection{Stage-1: Unimodal Self-Supervised Pretraining}
\label{subsec:per_modality_encoder}

Each of the physiological modalities is equipped with its own encoder, trained independently to capture modality-specific structure. Similar to NeuroNet's masked prediction combined with a contrastive learning setup \cite{lee2024neuronet}, we adopt a self-supervised learning (SSL) framework combining masked reconstruction, following the Masked Autoencoder (MAE) objective \cite{he2022masked}, and a contrastive learning objective to obtain robust and modality-invariant representations.

\subsubsection{Masked Reconstruction}
\label{sec:per_modality_MAE}
For each 30-second input signal $x$ of a given modality $m$, we partition the sequence into $P$ non-overlapping patches and randomly mask a fixed fraction $r$ (50\%) of them after adding token-wise positional embeddings.
The visible patches are processed by modality-specific Vision Transformers \cite{dosovitskiy2020image}, producing a latent sequence
$
    \mathbf{H_m} \in \mathbb{R}^{(1 - r)P \times D},
$
where $D$ is the embedding dimension. A lightweight decoder then reconstructs the full set of $P$ patches (both visible and masked) from this latent representation. After pretraining, this decoder is discarded, and the pretrained encoder serves as the feature extractor for downstream stages.

\subsubsection{Contrastive Framework}
To complement the reconstruction objective, we additionally train each encoder with a contrastive loss that encourages invariance across augmented views. For each signal $x$, we construct two views: the original signal and a lightly perturbed version of the same signal.

Both views are processed by the same encoder, and we obtain their CLS tokens, which are projected into a contrastive space and optimized using the NT-Xent loss \cite{chen2020simple}. This encourages representations from the same signal to align while separating different signals, yielding more discriminative and robust modality-specific embeddings.

\subsubsection{Combined Objective}
The full pretraining objective for each modality is the weighted sum of the MAE reconstruction loss and the contrastive loss:
\begin{equation}
    \mathcal{L}_{\text{total}}
    = \mathcal{L}_{\text{rec}}
    + \lambda_{\text{con}} \, \mathcal{L}_{\text{con}},
\end{equation}
where $\lambda_{\text{con}}$ balances the two terms. This combination encourages the encoders to learn features that best represent the input signals while remaining consistent across random perturbations. All modalities are pretrained independently under this objective, producing a set of modality-specific encoders $\{f_m\}$ used in the next cross-modal interaction stage. Additional training details of this process are included in Appendix \ref{app:unimodal_training}. 

\subsection{Stage-2: Cross-Modal Interaction and SSL}

\subsubsection{Random Modality Selection and Bimodal Stacking}
\label{sec:random_modality_selection}

Once the per-modality encoders have been pretrained, we construct bimodal inputs for the cross-attention module. At each training iteration (i.e. for every mini-batch), we randomly select two distinct modalities from the full set, and index individual samples within the mini-batch by
$i$. This sampling strategy ensures that the model is exposed to a wide range of modality combinations, promoting robustness to missing or noisy channels, common in clinical PSG recordings, and encouraging the model to learn generalizable cross-modal relationships.

Each selected modality is then encoded independently using its pretrained Stage-1 encoder $f_m$, producing patch-level embeddings.

This yields the final bimodal tensor
\[
    \mathbf{Z}_i \in \mathbb{R}^{2 \times P \times D},
\]
which serves as input to the bimodal cross-attention encoder.

\noindent\textbf{Relation to prior randomness-based methods.}
Randomness has previously been introduced into supervised sleep staging pipelines for robustness, but for fundamentally different objectives. For example, U-Sleep~\cite{UNet} randomly samples EEG and EOG channels to promote invariance to electrode placement, while wav2sleep~\cite{carter2024wav2sleep} applies random modality dropout to maintain performance under missing sensors.

In contrast, our approach randomly samples modality pairs from a diverse set of channels and explicitly trains cross-attention in a self-supervised setting. This design is intended to learn task-agnostic physiological relationships between modalities. Additional training details of Stage 2 are provided in Appendix \ref{app:cross_modal_training}.

\subsubsection{Global Time-Aware Conditioning}
\paragraph{Standard Positional Encodings} Before passing the bimodal embeddings $\mathbf{Z}_i$ to the cross-attention module, we enrich each patch with positional structure. Prior work has shown that positional encodings are essential for sleep-related transformers. In particular, \citet{shook2025stamp} and \citet{irani2502positional} evaluated combinations of spatial, temporal, and token encodings, with STAMP~\cite{shook2025stamp} demonstrating that using \emph{all three} components yields the best AUROC and Cohen's $\kappa$ on 3 of 4 EEG datasets. These findings suggest that richer positional structure improves physiological representation learning.

Motivated by this general insight, we incorporate the same triplet of learnable positional components. We denote by $\mathbf{z}_{i,j,p}$ the embedding of the $p$-th patch from
modality $j$ for the $i$-th sample in the mini-batch, and augment each patch as
\begin{equation}
    \mathbf{z}_{i,j,p}
    \leftarrow
    \mathbf{z}_{i,j,p}
    +
    \mathbf{s}_{j}
    +
    \mathbf{t}_{p}
    +
    \mathbf{n}_{j,p},
\end{equation}
where $\mathbf{s}_{j}$ is a spatial (modality-wise) embedding, $\mathbf{t}_{p}$ is a temporal (patch-wise) embedding, and $\mathbf{n}_{j,p}$ is a token-level embedding.

\paragraph{Global Time Conditioning}
While the above positional encodings capture \emph{local} temporal structure within each 30-second window, they do not reflect the \emph{global} progression of a full-night sleep session. Physiological patterns evolve over the night (REM density increases, respiratory stability shifts, and arousal probability decreases) yet such long-range trends are inaccessible to patch-level positional encodings. Our time conditioning instead encodes where an entire window occurs within its parent sleep session. This allows otherwise identical windows to be seen differently based on their position within the night. 

To incorporate this global context, we condition each example on its position within the sleep session. Each example corresponds to a fixed-length 30-second segment, and its position in the night can be described by the order in which it appears within the session. Let $\mathrm{se}_i$ denote this segment index for example $i$. To account for differences in session length across patients, we standardize this index using statistics of session length computed over the dataset, where $\mu_{\text{session}}$ and $\sigma_{\text{session}}$ denote the mean and standard deviation of the number of examples per sleep session:
\begin{equation}
    \hat{t}_i = \frac{\mathrm{se}_i - \mu_{\text{session}}}{\sigma_{\text{session}}}
\end{equation}

A lightweight two-layer network maps $\hat{t}_i$ to feature-wise scale and shift parameters $(\gamma_i,\beta_i) \in \mathbb{R}^{D}$. These vectors are layer-normalized and gated by learnable scalar coefficients initialized to zero, ensuring an identity transformation at initialization and a smooth learning trajectory.

Finally, we apply a time-dependent affine modulation after the spatial, temporal, and token encodings:
\begin{equation}
    \mathbf{z}_{i,j,p}
    \leftarrow
    \gamma_i \odot \mathbf{z}_{i,j,p} + \beta_i,
    \label{eq:time_film}
\end{equation}
introducing night-scale temporal structure that cannot be modeled by local positional encodings alone. While similar in form to FiLM-style conditioning~\cite{perez2018film}, our focus is on the conditioning signal—global time-of-sleep—rather than the specific conditioning mechanism.

\subsubsection{Bimodal Cross-Attention and SSL}

After enriching the bimodal tensor $\mathbf{Z}_i$ with the encodings from the previous section, we apply a second MAE-style masking step identical to the unimodal pretraining in Section~\ref{subsec:per_modality_encoder} and based on \citet{he2022masked}. We randomly mask 50\% of all patch tokens across both modalities and retain only the $V$ visible patches,
$
    \mathbf{Z}_i^{\text{vis}} \in \mathbb{R}^{2 \times V \times D}.
$

These visible tokens are processed by a bimodal cross-attention encoder, which performs two directional attention flows: each modality attends to the other’s remaining patches. This enables the model to capture physiologically meaningful relationships (e.g., SpO$_2$–respiratory coupling, EOG–EMG co-interactions). Because modality pairs are sampled randomly at every batch, the cross-attention module learns broad, dataset-agnostic structure across all modalities.
The output is a fused cross-modal representation,
$
    \tilde{\mathbf{Z}}_i \in \mathbb{R}^{2 \times V \times D},
$
which serves as input to the second-stage SSL objective.

Stage-2 pretraining uses the \emph{same} self-supervised objectives as the unimodal encoders (masked reconstruction~\cite{he2022masked} and NT-Xent contrastive learning~\cite{chen2020simple}) but now applied to cross-attended representations. This encourages the model to learn embeddings that are jointly reconstructive, meaningfully aligned, and highly transferable. While prior work such as~\cite{CASleepNet} has applied cross-attention in supervised sleep staging, these methods typically rely on only two modalities (e.g., EEG and EOG). In contrast, our framework integrates cross-attention within a self-supervised pipeline and supports a substantially broader set of physiological signals.

\paragraph{Gated Attention.}
Within the cross-attention block, we additionally apply a sigmoid-gated transformation after the scaled dot-product operation. Prior work (e.g., \cite{qiu2025gated}) has shown that such gating introduces beneficial non-linearities and sparsity while mitigating attention collapse. 

\paragraph{LoRA-based Fine-Tuning.}
During Stage-2, all modality-specific encoders are fine-tuned using Low-Rank Adaptation (LoRA) \cite{hu2022lora}. LoRA injects low-rank adapters into the attention projections while keeping the majority of parameters frozen, allowing efficient adaptation of the pretrained components without overfitting.

Together, these mechanisms yield rich representations that capture both modality-specific structure and robust inter-modal dependencies, forming a strong foundation for downstream sleep-related tasks.

\FloatBarrier
\begin{table*}[tb]
\centering
\caption{
Comparison of the non–time-aware baseline (BCNet) and BTCNet across six downstream tasks.
For each task, we report the best-performing modality pair selected using BCNet only.
Prevalence (\%Pos) is shown once per task for binary classification problems.
Results are averaged over three seeds and reported as mean (SD).
}
\label{tab:downstream_results}
\small
\setlength{\tabcolsep}{6pt}
\begin{tabular}{l c l ccc ccc}
\toprule
Task & \%Pos & Modality Pair &
\multicolumn{3}{c}{BCNet (Non Time-Aware)} &
\multicolumn{3}{c}{BTCNet } \\
\cmidrule(r){4-6} \cmidrule(r){7-9}
& & &
\shortstack{Acc\\(\%)} &
\shortstack{AUROC\\(\%)} &
\shortstack{F1\\(\%)} &
\shortstack{Acc\\(\%)} &
\shortstack{AUROC\\(\%)} &
\shortstack{F1\\(\%)} \\
\midrule

Oxygen Desaturation
& 8.78
& CAPNO -- SPO2
& \shortstack{88.40\\{\scriptsize (0.14)}}
& \shortstack{78.92\\{\scriptsize (0.13)}}
& \shortstack{38.91\\{\scriptsize (0.18)}}
& \textbf{\shortstack{89.92\\{\scriptsize (0.26)}}}
& \textbf{\shortstack{81.75\\{\scriptsize (0.15)}}}
& \textbf{\shortstack{44.80\\{\scriptsize (0.21)}}} \\

\midrule

Hypopnea
& 1.97
& EEG F4-M1 -- SPO2
& \shortstack{96.43\\{\scriptsize (0.05)}}
& \shortstack{82.96\\{\scriptsize (0.12)}}
& \shortstack{25.35\\{\scriptsize (0.22)}}
& \textbf{\shortstack{96.50\\{\scriptsize (0.24)}}}
& \textbf{\shortstack{86.13\\{\scriptsize (0.14)}}}
& \textbf{\shortstack{29.49\\{\scriptsize (0.10)}}} \\

\midrule

Apnea--Hypopnea
& 2.80
& EEG F3-M2 -- SPO2
& \shortstack{96.16\\{\scriptsize (0.02)}}
& \shortstack{82.83\\{\scriptsize (0.17)}}
& \shortstack{31.40\\{\scriptsize (0.57)}}
& \textbf{\shortstack{96.27\\{\scriptsize (0.10)}}}
& \textbf{\shortstack{86.04\\{\scriptsize (0.23)}}}
& \textbf{\shortstack{36.55\\{\scriptsize (0.04)}}} \\

\midrule

Apnea
& 0.83
& SPO2 -- EOG ROC-M1
& \shortstack{97.93\\{\scriptsize (0.06)}}
& \shortstack{82.16\\{\scriptsize (0.38)}}
& \shortstack{17.46\\{\scriptsize (0.66)}}
& \textbf{\shortstack{98.33\\{\scriptsize (0.06)}}}
& \textbf{\shortstack{85.11\\{\scriptsize (0.52)}}}
& \textbf{\shortstack{21.41\\{\scriptsize (1.09)}}} \\

\midrule

EEG-Arousal
& 4.71
& EOG LOC-M2 -- EMG CHIN1--CHIN2
& \shortstack{90.70\\{\scriptsize (0.30)}}
& \shortstack{78.53\\{\scriptsize (0.21)}}
& \shortstack{27.78\\{\scriptsize (0.23)}}
& \textbf{\shortstack{90.87\\{\scriptsize (0.32)}}}
& \textbf{\shortstack{81.82\\{\scriptsize (0.07)}}}
& \textbf{\shortstack{31.02\\{\scriptsize (0.25)}}} \\

\midrule

5-stage Sleep Scoring
& -
& EEG C3-M2 -- EOG ROC-M1
& \shortstack{57.12\\{\scriptsize (0.34)}}
& \shortstack{79.90\\{\scriptsize (0.01)}}
& \shortstack{55.08\\{\scriptsize (0.61)}}
& \textbf{\shortstack{60.98\\{\scriptsize (0.44)}}}
& \textbf{\shortstack{83.67\\{\scriptsize (0.05)}}}
& \textbf{\shortstack{59.83\\{\scriptsize (0.48)}}} \\

\bottomrule
\end{tabular}
\end{table*}

\section{Experiments and Results}
We demonstrate the effectiveness of BTCNet across multiple downstream tasks and datsets. Along with accuracy, AUROC and F1 score, event prevalence for each task is reported when appropriate to contextualize these metrics.

\subsection{Effectiveness of Global Time-Aware Pretraining}
\subsubsection{Time-Aware vs Non–Time-Aware Pretraining on NCH}
\label{sec:btcnet_vs_bcnet}

In this section, we evaluate BTCNet on multiple downstream sleep-related tasks on the NCH dataset and compare its performance to an otherwise identical variant that lacks global time-awareness, which we refer to as BCNet.

We first select informative modality pairs via a lightweight linear
probing sweep over all 120 possible channel pairs (see Appendix~\ref{app:modality_screening}). 
Importantly, modality selection is performed \emph{only} using the non--time-aware baseline (BCNet). BTCNet is then evaluated on the exact same modality combinations. Because these channels are optimized for the weaker baseline rather than for BTCNet, any observed improvements cannot be attributed to favorable channel selection and instead directly reflect the impact of incorporating global time-aware positional structure during pretraining.

For downstream evaluation, we freeze BTCNet and train linear classifiers on top of its representations. Specifically, we extract the modality-specific CLS tokens from BTCNet, concatenate them into a 1024-dimensional embedding, and use this representation as input to a linear classifier. We repeat the same procedure for BCNet. Unless stated otherwise, this linear probing setup, using frozen embeddings derived from concatenated CLS tokens, is used consistently for all BTCNet and BCNet evaluations throughout the paper. All downstream evaluations use a consistent 80/10/10 split across methods.

Table~\ref{tab:downstream_results} reports results for the single best-performing modality pair per task. Despite severe class imbalance across multiple binary tasks, both models exhibit strong linear separability, with BTCNet consistently achieving higher AUROC and F1 scores than BCNet across all six downstream tasks. Notably, these gains persist across all evaluated modality combinations in Appendix \ref{app:bcnet_vs_btcnet_additional_table}, indicating that the improvements are not driven by a single favorable channel pairing.

Because the encoders are frozen and the downstream models are linear, these improvements cannot be attributed to task-specific fine-tuning or increased model capacity. Instead, they reflect systematic differences in representation quality learned during pretraining. The consistent gains observed across diverse clinical tasks demonstrate that explicitly modeling global time structure leads to more robust, transferable, and clinically meaningful multimodal sleep representations.

\begin{table*}[tb]
\centering
\caption{
Linear probing performance on downstream tasks on the NCH test set. Binary tasks report Accuracy (Acc), AUROC, and F1. Sleep scoring reports Accuracy, weighted AUROC, and weighted F1. All metrics are expressed in \% and averaged across two seeds and reported as mean (SD)
}
\label{tab:linear_probe_comparison}
\small
\sisetup{
  detect-weight=true,
  detect-inline-weight=math,
  table-number-alignment=center
}
\setlength{\tabcolsep}{4pt}

\begin{tabular}{l l
                S[table-format=2.2] S[table-format=2.2] S[table-format=2.2]
                S[table-format=2.2] S[table-format=2.2] S[table-format=2.2]}
\toprule
Method & Metric &
{Oxygen Desaturation} & {Hypopnea} & {Apnea-Hypopnea} & {Apnea} & {EEG-Arousal} & {Sleep Scoring} \\
\midrule

\multirow{3}{*}{\makecell[l]{SleepFM (pool)}}
& Acc
& \multicolumn{1}{c}{\num{82.22}\,{\scriptsize(7.52)}}
& \multicolumn{1}{c}{{\bfseries\num{97.56}\,{\scriptsize(0.12)}}}
& \multicolumn{1}{c}{\num{95.70}\,{\scriptsize(1.58)}}
& \multicolumn{1}{c}{\num{96.43}\,{\scriptsize(1.76)}}
& \multicolumn{1}{c}{{\bfseries\num{94.88}\,{\scriptsize(0.51)}}}
& \multicolumn{1}{c}{{\bfseries\num{71.95}\,{\scriptsize(0.25)}}} \\

& AUROC
& \multicolumn{1}{c}{\num{79.72}\,{\scriptsize(0.41)}}
& \multicolumn{1}{c}{\num{85.36}\,{\scriptsize(0.17)}}
& \multicolumn{1}{c}{\num{85.42}\,{\scriptsize(0.00)}}
& \multicolumn{1}{c}{{\bfseries\num{87.88}\,{\scriptsize(0.10)}}}
& \multicolumn{1}{c}{{\bfseries\num{89.51}\,{\scriptsize(0.04)}}}
& \multicolumn{1}{c}{{\bfseries\num{92.00}\,{\scriptsize(0.22)}}} \\

& F1
& \multicolumn{1}{c}{\num{32.94}\,{\scriptsize(0.01)}}
& \multicolumn{1}{c}{\num{17.18}\,{\scriptsize(1.72)}}
& \multicolumn{1}{c}{\num{14.66}\,{\scriptsize(12.38)}}
& \multicolumn{1}{c}{\num{15.44}\,{\scriptsize(2.17)}}
& \multicolumn{1}{c}{\num{28.02}\,{\scriptsize(14.79)}}
& \multicolumn{1}{c}{{\bfseries\num{73.09}\,{\scriptsize(0.20)}}} \\

\midrule

\multirow{3}{*}{\makecell[l]{SleepFM (no pool)}}
& Acc
& \multicolumn{1}{c}{\num{86.56}\,{\scriptsize(3.18)}}
& \multicolumn{1}{c}{\num{92.50}\,{\scriptsize(2.21)}}
& \multicolumn{1}{c}{\num{85.23}\,{\scriptsize(2.66)}}
& \multicolumn{1}{c}{\num{87.73}\,{\scriptsize(10.38)}}
& \multicolumn{1}{c}{\num{94.67}\,{\scriptsize(0.30)}}
& \multicolumn{1}{c}{\num{70.93}\,{\scriptsize(1.27)}} \\

& AUROC
& \multicolumn{1}{c}{\num{78.92}\,{\scriptsize(0.39)}}
& \multicolumn{1}{c}{\num{84.25}\,{\scriptsize(1.11)}}
& \multicolumn{1}{c}{\num{83.53}\,{\scriptsize(1.04)}}
& \multicolumn{1}{c}{\num{86.49}\,{\scriptsize(0.77)}}
& \multicolumn{1}{c}{\num{88.08}\,{\scriptsize(1.47)}}
& \multicolumn{1}{c}{\num{91.38}\,{\scriptsize(0.40)}} \\

& F1
& \multicolumn{1}{c}{\num{34.19}\,{\scriptsize(1.26)}}
& \multicolumn{1}{c}{\num{19.55}\,{\scriptsize(2.02)}}
& \multicolumn{1}{c}{\num{18.80}\,{\scriptsize(1.47)}}
& \multicolumn{1}{c}{\num{10.86}\,{\scriptsize(5.20)}}
& \multicolumn{1}{c}{{\bfseries\num{36.91}\,{\scriptsize(5.89)}}}
& \multicolumn{1}{c}{\num{70.97}\,{\scriptsize(1.93)}} \\
\midrule

\multirow{3}{*}{\makecell[l]{{PedSleepMAE}}}
& Acc
& \multicolumn{1}{c}{\num{84.08}\,{\scriptsize(0.00)}}
& \multicolumn{1}{c}{\num{95.07}\,{\scriptsize(0.15)}}
& \multicolumn{1}{c}{\num{94.05}\,{\scriptsize(0.51)}}
& \multicolumn{1}{c}{\num{97.47}\,{\scriptsize(0.11)}}
& \multicolumn{1}{c}{\num{90.34}\,{\scriptsize(0.01)}}
& \multicolumn{1}{c}{\num{71.88}\,{\scriptsize(1.60)}} \\

& AUROC
& \multicolumn{1}{c}{\num{77.18}\,{\scriptsize(0.04)}}
& \multicolumn{1}{c}{\num{79.47}\,{\scriptsize(0.08)}}
& \multicolumn{1}{c}{\num{79.52}\,{\scriptsize(0.20)}}
& \multicolumn{1}{c}{\num{79.62}\,{\scriptsize(0.71)}}
& \multicolumn{1}{c}{\num{79.81}\,{\scriptsize(0.15)}}
& \multicolumn{1}{c}{\num{90.82}\,{\scriptsize(0.22)}} \\

& F1
& \multicolumn{1}{c}{\num{32.48}\,{\scriptsize(0.10)}}
& \multicolumn{1}{c}{\num{16.79}\,{\scriptsize(0.26)}}
& \multicolumn{1}{c}{\num{20.67}\,{\scriptsize(0.50)}}
& \multicolumn{1}{c}{\num{10.83}\,{\scriptsize(0.25)}}
& \multicolumn{1}{c}{\num{28.88}\,{\scriptsize(0.61)}}
& \multicolumn{1}{c}{\num{71.93}\,{\scriptsize(0.65)}} \\

\midrule

\multirow{3}{*}{\makecell[l]{BTCNet (ours)}}
& Acc
& \multicolumn{1}{c}{{\bfseries\num{89.97}\,{\scriptsize(0.13)}}}
& \multicolumn{1}{c}{\num{96.93}\,{\scriptsize(0.01)}}
& \multicolumn{1}{c}{{\bfseries\num{96.36}\,{\scriptsize(0.21)}}}
& \multicolumn{1}{c}{{\bfseries\num{98.16}\,{\scriptsize(0.12)}}}
& \multicolumn{1}{c}{\num{90.78}\,{\scriptsize(0.28)}}
& \multicolumn{1}{c}{\num{61.98}\,{\scriptsize(0.39)}} \\

& AUROC
& \multicolumn{1}{c}{{\bfseries\num{81.95}\,{\scriptsize(0.14)}}}
& \multicolumn{1}{c}{{\bfseries\num{86.61}\,{\scriptsize(0.08)}}}
& \multicolumn{1}{c}{{\bfseries\num{86.03}\,{\scriptsize(0.19)}}}
& \multicolumn{1}{c}{\num{85.03}\,{\scriptsize(0.62)}}
& \multicolumn{1}{c}{\num{81.16}\,{\scriptsize(0.13)}}
& \multicolumn{1}{c}{\num{84.52}\,{\scriptsize(0.01)}} \\

& F1
& \multicolumn{1}{c}{{\bfseries\num{44.95}\,{\scriptsize(0.52)}}}
& \multicolumn{1}{c}{{\bfseries\num{29.88}\,{\scriptsize(0.68)}}}
& \multicolumn{1}{c}{{\bfseries\num{37.07}\,{\scriptsize(0.43)}}}
& \multicolumn{1}{c}{{\bfseries\num{20.59}\,{\scriptsize(0.88)}}}
& \multicolumn{1}{c}{\num{31.27}\,{\scriptsize(0.17)}}
& \multicolumn{1}{c}{\num{61.55}\,{\scriptsize(0.41)}} \\
\bottomrule
\end{tabular}

\vspace{1mm}
\footnotesize{
BTCNet uses task-specific 2-channel pairs that were one of the top combinations on NCH:
Desaturation (CAPNO--SPO2),
Hypopnea (EEG O1-M2--SPO2),
Apnea-Hypopnea (EEG-O1-M2--SPO2),
Apnea (EEG F3-M2--SPO2),
EEG-Arousal (EEG O1-M2--EMG CHIN1--CHIN2),
Sleep Scoring (EEG F3-M2--EOG ROC-M1).
}
\end{table*}

\begin{table*}[tb]
\centering
\caption{
Cross-dataset inference performance when training on NCH and evaluating directly on the independent CHAT test set.
Binary tasks report Accuracy (Acc), AUROC, and F1.
Sleep scoring reports Accuracy, weighted AUROC, and weighted F1. All metrics are expressed in \% and averaged across two seeds and reported as mean (SD).}
\label{tab:cross_dataset_inference}
\small
\sisetup{
  detect-weight=true,
  detect-inline-weight=math,
  table-number-alignment=center
}
\setlength{\tabcolsep}{4pt}

\begin{tabular}{l l
                S[table-format=2.2] S[table-format=2.2] S[table-format=2.2]
                S[table-format=2.2] S[table-format=2.2] S[table-format=2.2]}
\toprule
Method & Metric &
{Oxygen Desaturation} & {Hypopnea} & {Apnea-Hypopnea} & {Apnea} & {EEG-Arousal} & {Sleep Scoring} \\
\midrule

\multirow{3}{*}{\makecell[l]{SleepFM (pool)}}
& Acc
& \multicolumn{1}{c}{\num{77.95}\,{\scriptsize(3.63)}}
& \multicolumn{1}{c}{\num{90.97}\,{\scriptsize(1.66)}}
& \multicolumn{1}{c}{\num{93.04}\,{\scriptsize(1.20)}}
& \multicolumn{1}{c}{\num{94.77}\,{\scriptsize(1.37)}}
& \multicolumn{1}{c}{\num{93.32}\,{\scriptsize(0.71)}}
& \multicolumn{1}{c}{\bfseries\num{78.59}\,{\scriptsize(1.93)}} \\

& AUROC
& \multicolumn{1}{c}{\num{84.56}\,{\scriptsize(0.01)}}
& \multicolumn{1}{c}{\num{86.76}\,{\scriptsize(0.05)}}
& \multicolumn{1}{c}{\num{87.49}\,{\scriptsize(0.14)}}
& \multicolumn{1}{c}{\bfseries\num{85.86}\,{\scriptsize(0.50)}}
& \multicolumn{1}{c}{\bfseries\num{94.39}\,{\scriptsize(0.06)}}
& \multicolumn{1}{c}{\bfseries\num{96.64}\,{\scriptsize(0.06)}} \\

& F1
& \multicolumn{1}{c}{\num{52.38}\,{\scriptsize(1.57)}}
& \multicolumn{1}{c}{\num{29.85}\,{\scriptsize(1.21)}}
& \multicolumn{1}{c}{\num{28.74}\,{\scriptsize(10.96)}}
& \multicolumn{1}{c}{\num{21.47}\,{\scriptsize(0.41)}}
& \multicolumn{1}{c}{\bfseries\num{63.30}\,{\scriptsize(1.07)}}
& \multicolumn{1}{c}{\bfseries\num{79.62}\,{\scriptsize(1.85)}} \\

\midrule

\multirow{3}{*}{\makecell[l]{SleepFM (no pool)}}
& Acc
& \multicolumn{1}{c}{\num{82.73}\,{\scriptsize(0.56)}}
& \multicolumn{1}{c}{\num{72.95}\,{\scriptsize(20.07)}}
& \multicolumn{1}{c}{\num{81.56}\,{\scriptsize(7.40)}}
& \multicolumn{1}{c}{\num{82.41}\,{\scriptsize(12.63)}}
& \multicolumn{1}{c}{\bfseries\num{94.25}\,{\scriptsize(0.17)}}
& \multicolumn{1}{c}{\num{70.93}\,{\scriptsize(6.37)}} \\

& AUROC
& \multicolumn{1}{c}{\num{84.72}\,{\scriptsize(0.12)}}
& \multicolumn{1}{c}{\num{85.94}\,{\scriptsize(0.72)}}
& \multicolumn{1}{c}{\num{86.99}\,{\scriptsize(0.25)}}
& \multicolumn{1}{c}{\num{85.37}\,{\scriptsize(0.24)}}
& \multicolumn{1}{c}{\num{94.21}\,{\scriptsize(0.09)}}
& \multicolumn{1}{c}{\num{95.88}\,{\scriptsize(0.15)}} \\

& F1
& \multicolumn{1}{c}{\num{54.62}\,{\scriptsize(0.62)}}
& \multicolumn{1}{c}{\num{22.30}\,{\scriptsize(8.92)}}
& \multicolumn{1}{c}{\num{33.31}\,{\scriptsize(5.68)}}
& \multicolumn{1}{c}{\num{18.06}\,{\scriptsize(6.15)}}
& \multicolumn{1}{c}{\num{60.75}\,{\scriptsize(3.38)}}
& \multicolumn{1}{c}{\num{72.40}\,{\scriptsize(5.70)}} \\

\midrule

\multirow{3}{*}{\makecell[l]{SleepFM (bimodal)}}
& Acc
& \multicolumn{1}{c}{\num{83.30}\,{\scriptsize(1.27)}}
& \multicolumn{1}{c}{\bfseries\num{94.87}\,{\scriptsize(1.07)}}
& \multicolumn{1}{c}{\num{93.59}\,{\scriptsize(0.52)}}
& \multicolumn{1}{c}{\bfseries\num{95.34}\,{\scriptsize(0.20)}}
& \multicolumn{1}{c}{\num{90.61}\,{\scriptsize(0.16)}}
& \multicolumn{1}{c}{\num{76.05}\,{\scriptsize(2.05)}} \\

& AUROC
& \multicolumn{1}{c}{\num{86.39}\,{\scriptsize(0.14)}}
& \multicolumn{1}{c}{\num{86.42}\,{\scriptsize(0.38)}}
& \multicolumn{1}{c}{\num{86.43}\,{\scriptsize(0.13)}}
& \multicolumn{1}{c}{\num{84.27}\,{\scriptsize(0.09)}}
& \multicolumn{1}{c}{\num{91.91}\,{\scriptsize(0.14)}}
& \multicolumn{1}{c}{\num{95.67}\,{\scriptsize(0.14)}} \\

& F1
& \multicolumn{1}{c}{\num{57.61}\,{\scriptsize(0.68)}}
& \multicolumn{1}{c}{\num{29.91}\,{\scriptsize(5.80)}}
& \multicolumn{1}{c}{\num{39.25}\,{\scriptsize(3.68)}}
& \multicolumn{1}{c}{\num{23.43}\,{\scriptsize(0.85)}}
& \multicolumn{1}{c}{\num{54.50}\,{\scriptsize(0.07)}}
& \multicolumn{1}{c}{\num{77.55}\,{\scriptsize(1.25)}} \\

\midrule

\multirow{3}{*}{\makecell[l]{BTCNet (ours)}}
& Acc
& \multicolumn{1}{c}{\bfseries\num{89.93}\,{\scriptsize(0.09)}}
& \multicolumn{1}{c}{\num{94.78}\,{\scriptsize(0.35)}}
& \multicolumn{1}{c}{\bfseries\num{94.15}\,{\scriptsize(0.01)}}
& \multicolumn{1}{c}{\num{95.15}\,{\scriptsize(0.01)}}
& \multicolumn{1}{c}{\num{89.78}\,{\scriptsize(0.06)}}
& \multicolumn{1}{c}{\num{55.01}\,{\scriptsize(0.02)}} \\

& AUROC
& \multicolumn{1}{c}{\bfseries\num{90.65}\,{\scriptsize(0.04)}}
& \multicolumn{1}{c}{\bfseries\num{88.00}\,{\scriptsize(0.55)}}
& \multicolumn{1}{c}{\bfseries\num{88.32}\,{\scriptsize(0.29)}}
& \multicolumn{1}{c}{\num{84.64}\,{\scriptsize(0.37)}}
& \multicolumn{1}{c}{\num{84.93}\,{\scriptsize(0.31)}}
& \multicolumn{1}{c}{\num{84.62}\,{\scriptsize(0.04)}} \\

& F1
& \multicolumn{1}{c}{\bfseries\num{68.01}\,{\scriptsize(0.25)}}
& \multicolumn{1}{c}{\bfseries\num{42.42}\,{\scriptsize(0.85)}}
& \multicolumn{1}{c}{\bfseries\num{49.89}\,{\scriptsize(0.47)}}
& \multicolumn{1}{c}{\bfseries\num{27.82}\,{\scriptsize(0.65)}}
& \multicolumn{1}{c}{\num{44.86}\,{\scriptsize(0.37)}}
& \multicolumn{1}{c}{\num{54.88}\,{\scriptsize(0.46)}} \\
\bottomrule
\end{tabular}

\vspace{1mm}
\footnotesize{
BTCNet and SleepFM (bimodal) use task-specific two-channel pairs:
Desaturation (CAPNO--SPO2),
Hypopnea (EEG O1-M2--SPO2),
Apnea--Hypopnea (EEG O1-M2--SPO2),
Apnea (EEG F3-M2--SPO2),
EEG-Arousal (EEG O1-M2--EMG CHIN1--CHIN2),
and Sleep Scoring (EEG F3-M2--EEG F4-M1), where EEG F4-M1 replaces EOG ROC-M1 on CHAT due to its absence (both are BAS channels).
}
\end{table*}

\subsubsection{Generalization of Time-Aware Pretraining to CHAT}

We evaluate the generalization of our global time-aware
pretraining approach on the unseen CHAT dataset.

Both BTCNet (time-aware) and BCNet (non–time-aware), pretrained
on NCH, are fine-tuned on CHAT using LoRA adapters applied
to the attention layers of the unimodal encoders and the
cross-modal fusion module. Following the same evaluation protocol as in Section~\ref{sec:btcnet_vs_bcnet}, we first identify the best-performing modality pair using the fine-tuned non–time-aware baseline (BCNet). We then evaluate both BCNet and BTCNet on this fixed channel configuration. This controlled setup isolates the contribution of global time-aware modeling while holding both the dataset and channel selection procedure constant. Additional details are provided in Appendix~\ref{app:finetuning_CHAT}.

Table~\ref{tab:chat_finetune_time_vs_notime} in Appendix~\ref{app:finetuning_CHAT} reports the results of this comparison. Across all clinical tasks, BTCNet consistently outperforms BCNet after fine-tuning on CHAT, despite modality selection being optimized for the non–time-aware baseline. While absolute accuracy remains high for both models due to class imbalance, BTCNet achieves substantially higher AUROC and F1 scores, indicating improved class separability. 

These results demonstrate that the benefits of global time-aware pretraining persist under dataset shift and are not specific to the NCH cohort. Together with the NCH results, this provides strong evidence that incorporating global time-of-sleep information during pretraining leads to more robust and transferable sleep representations.

\subsection{Comparison Against Existing SSL Baselines}
\subsubsection{Linear Probing on NCH (In-Dataset Evaluation)}
\label{sec:probe_NCH}
We next compare BTCNet against relevant and existing multimodal self-supervised learning (SSL) baselines for sleep representation learning whose checkpoints are publicly available and perform tasks beyond sleep staging: SleepFM and PedSleepMAE. 

To ensure a fair and consistent evaluation, we follow the same frozen-backbone, linear-probe protocol described in Section \ref{sec:btcnet_vs_bcnet} for all methods. For identifying the two most optimal modalities for BTCNet, we follow the same fast logistic regression method described in Appendix ~\ref{app:modality_screening} for each task. For SleepFM, we follow its original input specification and utilize all available channels in the NCH dataset, grouped into BAS (brain activity signals), RESP (respiratory signals), and EMG modality categories. Because SleepFM does not report a specific linear probing setup, we evaluate multiple reasonable embedding aggregation strategies—i) flattening raw embeddings directly without any pooling and ii) average pooling across time followed by flattening—and report both results for fairness. For PedSleepMAE, we follow the authors’ original evaluation protocol and use their provided linear probing implementation without modification, as the same dataset and channel configuration are used.

According to Table~\ref{tab:linear_probe_comparison}, BTCNet consistently outperforms SleepFM and PedSleepMAE on respiratory-related tasks and achieves comparable performance on EEG-Arousal and sleep staging. These results suggest that BTCNet learns robust and transferable representations that generalize well across diverse clinical sleep tasks, despite differences in task characteristics and signal types. Overall, this comparison highlights the effectiveness of BTCNet’s learned representations under a frozen-backbone, linear-probing evaluation, indicating strong representation quality rather than task-specific architectural advantages. Additional training details are included in Appendix \ref{app:tab:linear_probing}.

\subsubsection{Cross-Dataset Linear Probing on CHAT}
In this section, we evaluate our models on the unseen CHAT dataset. Since this dataset was not used for pretraining by any of our models, the dataset serves as a strong choice for validating our results on an external site. We compare BTCNet (pretrained on NCH) against the two SleepFM variants described in Section \ref{sec:probe_NCH} in this section and report the results in Table \ref{tab:cross_dataset_inference}. Additionally, to enable a more direct comparison, we evaluate
SleepFM using the same two task-specific modalities selected
for BTCNet, with temporal average pooling applied to the
resulting representations, noted as SleepFM (bimodal).

As shown in Table~\ref{tab:cross_dataset_inference}, BTCNet demonstrates strong cross-dataset generalization when trained on NCH and evaluated directly on the independent CHAT cohort. In particular, BTCNet substantially outperforms all SleepFM variants on respiratory-related tasks, including oxygen desaturation, hypopnea, and apnea–hypopnea, achieving consistently higher F1 and AUROC scores despite severe class imbalance. In contrast, SleepFM variants outperform BTCNet on EEG-dominant tasks such as EEG-Arousal and sleep staging.

Among the SleepFM baselines, the bimodal variant consistently performs best on respiratory tasks, outperforming both pooled and all-modality variants and remaining competitive on arousal and sleep scoring. This suggests that selecting task-specific modality pairs can be more informative than indiscriminately incorporating all available channels. Notably, these informative modality pairings are identified using BTCNet’s cross-modal attention during pretraining. While further analysis is required to establish causality, this result suggests that modality relevance learned by BTCNet may transfer to other models, such as SleepFM, when used for downstream task configuration.

\section{Discussion and Conclusion}

This work examined whether explicitly modeling \emph{global time-of-sleep context} during multimodal self-supervised pretraining improves representation quality for pediatric sleep analysis. Across all controlled experiments, our results consistently show that global time-aware conditioning provides a meaningful and transferable inductive bias.

On the NCH dataset, BTCNet systematically outperforms an otherwise identical non–time-aware baseline (BCNet) across all six downstream tasks. Because modality selection is performed exclusively using BCNet and all evaluations rely on frozen encoders with linear classifiers, these gains cannot be attributed to favorable channel choice, increased model capacity, or task-specific adaptation. Instead, they reflect differences in representation quality learned during pretraining. The largest improvements are observed for respiratory-related tasks, such as oxygen desaturation, hypopnea, and apnea detection, which exhibit strong night-scale temporal structure across sleep cycles.

For EEG-dominant tasks including EEG-Arousal and sleep staging, BTCNet remains competitive with existing multimodal SSL baselines, despite being evaluated using only bimodal inputs. This suggests that incorporating global temporal context does not compromise performance on EEG-centric tasks while substantially improving performance on respiration-driven modalities.

Importantly, the benefits of time-aware pretraining persist under dataset shift. When evaluated on the independent CHAT cohort, both via fine-tuning and direct cross-dataset linear probing, BTCNet consistently outperforms non–time-aware baselines and existing multimodal SSL models on respiratory tasks, which demonstrates its robust cross-dataset generalization. An additional observation is that SleepFM performs best on respiratory tasks when using the modality pairs identified by BTCNet rather than all modalities, though this may be coincidental. Lastly, we emphasize that global time is not intended as a standalone predictive signal, but rather as a contextual conditioning mechanism that complements multimodal physiological representations during pretraining. 

\paragraph{Future work} While this work focuses on a hybrid masked-reconstruction and contrastive framework, an important future direction is to examine whether similar benefits could occur in purely contrastive SSL setups and other alternative architectures. Additionally, our linear probing experiments suggest that other temporal heads, such as lightweight sequence models fitted on top of frozen representations, may further improve downstream performance without requiring full end-to-end retraining. All in all, these directions point toward a broader role for global temporal modeling in scalable and generalizable sleep representation learning.

\section*{Acknowledgement}
This work was partially supported by the AI Acceleration Program at the University of North Carolina at Chapel Hill.

The Childhood Adenotonsillectomy Trial (CHAT) was supported by the National Institutes of Health (HL083075, HL083129, UL1-RR-024134, UL1 RR024989). The National Sleep Research Resource was supported by the National Heart, Lung, and Blood Institute (R24 HL114473, 75N92019R002).

\nocite{langley00}

\bibliography{paper}
\bibliographystyle{abbrvnat}

\newpage
\appendix
\onecolumn

\section{Datasets}
\label{app:nch_dataset}
In this section, we provide more information about our datasets. 

\subsection{Nationwide Children's Hospital (NCH) Dataset}
\paragraph{Selected Channels.}
From the NCH dataset \cite{PhysioNet-nch-sleep-3.1.0, goldberger2000physiobank,lee2022large}, we select the 16 most commonly available channels, spanning the following physiological categories:
\begin{itemize}
    \item \textbf{Electroencephalography (EEG, 7):}
    C3--M2, O1--M2, O2--M1, CZ--O1, C4--M1, F4--M1, F3--M2
    \item \textbf{Electrooculography (EOG, 2):}
    LOC--M2, ROC--M1
    \item \textbf{Electromyography (EMG, 1):}
    CHIN1--CHIN2
    \item \textbf{Respiratory signals (3):}
    thoracic effort, abdominal effort, CPAP airflow (C-FLOW)
    \item \textbf{Gas exchange (2):}
    oxygen saturation (SpO$_2$), end-tidal CO$_2$ (CAPNO)
    \item \textbf{Snoring (1):}
    SNORE
\end{itemize}

\paragraph{Preprocessing.} All signals are downsampled to 128\,Hz and normalized to zero mean and unit variance for stable training.

\paragraph{Dataset Splits.}
For downstream tasks, the dataset is partitioned into training, validation, and test splits (80-10-10).
Each split consists of fixed-length examples.
We report the total number of examples in each split in Table \ref{tab:dataset_split_examples}.

\begin{table}[b]
\centering
\caption{Number of examples in each dataset split for NCH and CHAT.}
\label{tab:dataset_split_examples}
\small
\setlength{\tabcolsep}{6pt}
\begin{tabular}{lcc}
\toprule
\textbf{Split} & \textbf{NCH} & \textbf{CHAT} \\
\midrule
Train      & 1{,}865{,}984 & 412{,}800 \\
Validation &   233{,}216 &  51{,}584 \\
Test       &   233{,}344 &  51{,}712 \\
\midrule
\textbf{Total} & 2{,}332{,}544 & 516{,}096 \\
\bottomrule
\end{tabular}
\end{table}

\paragraph{Class Imbalance and Loss Weights.}
Several downstream tasks derived from the NCH dataset exhibit substantial class imbalance, particularly for rare clinical events such as apnea, hypopnea, and oxygen desaturation.
To mitigate bias toward majority classes during supervised training, we apply task-specific class weighting in the loss function.

Table~\ref{tab:class_weights_combined} summarizes the class weights used for each downstream task.
For binary tasks, weights are specified for the negative (0) and positive (1) classes.
For sleep staging, class weights are provided for all five sleep stages.

\begin{table}[tb]
\centering
\caption{Class weights used for downstream evaluations on NCH and CHAT.}
\label{tab:class_weights_combined}
\small
\setlength{\tabcolsep}{5pt}
\begin{tabular}{l c cc}
\toprule
\textbf{Task} & \textbf{Class} & \textbf{NCH} & \textbf{CHAT} \\
\midrule
Apnea &
0 / 1 &
1.0 / 121.0 &
1.02 / 41.64 \\
\addlinespace

Hypopnea &
0 / 1 &
1.0 / 50.0 &
1.04 / 26.45 \\
\addlinespace

Apnea--Hypopnea &
0 / 1 &
1.0 / 37.0 &
1.06 / 16.88 \\
\addlinespace

EEG Arousal &
0 / 1 &
0.52 / 10.66 &
1.09 / 12.39 \\
\addlinespace

Oxygen Desaturation &
0 / 1 &
0.55 / 5.67 &
1.20 / 5.99 \\
\addlinespace

Sleep Scoring &
0--4 &
0.9 / 5.0 / 0.9 / 0.9 / 0.9 &
3.92 / 15.65 / 3.28 / 4.22 / 7.20 \\
\bottomrule
\end{tabular}
\end{table}

\subsection{Childhood Adenotonsillectomy Trial (CHAT)}
\label{app:chat_dataset}
\paragraph{Selected Channels.}
For CHAT \cite{marcus2013randomized, zhang2018national}, we utilize its 12 modalities that are also present in NCH or share a similar physiological function. For example, SaO$_2$ serves as the closest alternative to SpO$_2$, as both measure oxygen saturation. Channels without a clear functional match are excluded from analysis.

\paragraph{Preprocessing.} Similar to NCH, all signals are downsampled to 128\,Hz and normalized to zero mean and unit variance for stable training.

\paragraph{Dataset Splits.} We keep the same split as NCH and report the total number of examples in each split in Table \ref{tab:dataset_split_examples}.

\paragraph {Class Imbalance and Loss Weights} Similar to NCH, we report the class weights that were used when using the CHAT dataset for downstream evaluation tasks in Table \ref{tab:class_weights_combined}.

\section{Pretraining BTCNet}
\subsection{Stage 1: Unimodal Encoder Training}
\label{app:unimodal_training}
\paragraph{Model Architecture.} We list our MAE architecture's parameters in Table~\ref{tab:architecture_details} trained separately for each of our modalities.

\begin{table}[tb]
\centering
\caption{Model architecture details for unimodal encoders.}
\label{tab:architecture_details}
\small
\begin{tabular}{l c}
\toprule
\textbf{Component} & \textbf{Value} \\
\midrule
Patch Size & 8 time samples \\
Mask Ratio & 50\% \\
Encoder embedding dimension & 512 \\
Encoder layers & 6 \\
Encoder attention heads & 8 \\
Decoder embedding dimension & 512 \\
Decoder layers & 4 \\
Decoder attention heads & 4 \\
\bottomrule
\end{tabular}
\end{table}

\paragraph{Training Setup.} Table \ref{tab:training_setup} lists the hyperparameters used in this experiment. For computational efficiency, each training epoch is defined as a fixed number of iterations rather than a full pass over the dataset. Under this setup, approximately eight training epochs correspond to one full pass over the training data. Because epochs are defined using a fixed iteration budget, 24 warm-up epochs and a patience of 100 epochs correspond to approximately 3 and 12 full passes over the training data, respectively. All experiments follow the same overall training protocol, with mild input noise applied only for non-respiratory channels (EEG, EOG, EMG) and standard stabilization choices used where necessary to ensure stable training.

\paragraph{Reconstruction and Contrastive Objectives.}
BTCNet is trained using a combination of masked reconstruction and contrastive learning losses.
Let $x$ denote the input signal segmented into patches.

\paragraph{Reconstruction Loss.}
We use a masked mean-squared error (MSE) reconstruction loss computed only over masked signal positions:
\begin{equation}
\mathcal{L}_{\text{recon}}
=
\frac{1}{|M|}
\sum_{(c,t)\in M}
\left\| \hat{x}_{c,t} - x_{c,t} \right\|_2^2 ,
\end{equation}
where $x_{c,t}$ and $\hat{x}_{c,t}$ denote the ground-truth and reconstructed signal values for channel $c$ at time index $t$, and $M$ denotes the set of masked positions. For unimodal encoder training, this formulation reduces to just a one-channel case.

\paragraph{Contrastive Loss (NT-Xent).}
Given a batch of $N$ paired representations $\{(z_i^{(1)}, z_i^{(2)})\}_{i=1}^{N}$ obtained from two augmented views, we construct a set of $2N$ $\ell_2$-normalized embeddings $\{z_k\}_{k=1}^{2N}$ by concatenation.
Cosine similarity is used as the similarity function.

For a positive pair $(i,j)$ corresponding to two views of the same example, the NT-Xent loss is defined as
\begin{equation}
\ell_{i,j}
=
-\log
\frac{
\exp\!\left(\mathrm{sim}(z_i, z_j) / \tau \right)
}{
\sum_{k=1}^{2N}
\mathbf{1}\{k \neq i\}
\exp\!\left(\mathrm{sim}(z_i, z_k) / \tau \right)
},
\end{equation}
where $\mathbf{1}\{\cdot\}$ denotes the indicator function, $\mathrm{sim}(\cdot,\cdot)$ is cosine similarity, and $\tau$ is a temperature parameter.
The final contrastive objective averages this loss over all positive pairs in the batch.

\begin{table}[tb]
\centering
\caption{Training configuration and optimization details for unimodal encoders.}
\label{tab:training_setup}
\small
\setlength{\tabcolsep}{6pt}
\begin{tabular}{l c}
\toprule
\textbf{Parameter} & \textbf{Value} \\
\midrule
Batch size & 128 \\
Iterations per epoch & 2000 \\
Maximum epochs & 850 \\
Warm-up epochs & 24 \\
Early stopping patience & 100 epochs \\
Optimizer & Adam \\ 
Learning rate & $1 \times 10^{-4}$ \\
Weight decay & $1 \times 10^{-5}$ \\
Contrastive loss weight ($\lambda_{\text{contrast}}$) & Linearly ramped to 1.0 \\
Learning rate schedule & Linear warm-up followed by cosine decay \\
\bottomrule
\end{tabular}
\end{table}

\subsection{Stage 2: Cross-Modal Interaction and SSL} \label{app:cross_modal_training}

\paragraph{Model Architecture} We enlist the parameters used for our MAE architecture in Stage 2 in Table \ref{tab:crossmodal_arch}. 

\begin{table}[tb]
\centering
\caption{Stage-2 cross-modal model architecture and pretraining configuration.}
\label{tab:crossmodal_arch}
\small
\setlength{\tabcolsep}{6pt}
\begin{tabular}{l c}
\toprule
\textbf{Component} & \textbf{Value} \\
\midrule
Patch size & 8 samples \\
Mask ratio & 50\% \\
Encoder embedding dimension & 512 \\
Encoder layers & 10 \\
Encoder attention heads & 8 \\
Decoder embedding dimension & 512 \\
Decoder layers & 4 \\
Decoder attention heads & 4 \\
\bottomrule
\end{tabular}
\end{table}

\paragraph{Training Setup.} During Stage-2 training, pairs of modalities are randomly sampled at each iteration and fused using a cross-attention module conditioned on global time information. Unimodal encoders pretrained in Stage-1 are frozen, with lightweight LoRA adapters applied to their attention layers. We list the hyperparameters used in this process in Table \ref{tab:crossmodal_train}. Similar to the setup described in Appendix \ref{app:unimodal_training}, approximately eight training epochs correspond to one full pass over the training set.

\paragraph{LoRA Adaptation Details.} For parameter-efficient fine-tuning, we apply Low-Rank Adaptation (LoRA) to the attention layers of the pretrained unimodal encoders during Stage-2 training and downstream fine-tuning. Specifically, LoRA adapters are inserted into the query–key–value and output projection matrices of the self-attention blocks. We use a low-rank configuration with rank $r=8$, scaling factor $\alpha=16$, and dropout rate $0.05$. All original encoder parameters remain frozen, and only the LoRA parameters are updated during training.

\begin{table}[tb]
\centering
\caption{Stage-2 cross-modal training configuration.}
\label{tab:crossmodal_train}
\small
\setlength{\tabcolsep}{6pt}
\begin{tabular}{l c}
\toprule
\textbf{Parameter} & \textbf{Value} \\
\midrule
Batch size & 64 \\
Iterations per epoch & 4000 \\
Epochs & 200 \\
Warm-up epochs & 24 \\
Optimizer & Adam \\
Learning rate & $1 \times 10^{-4}$ \\
Weight decay & $1 \times 10^{-5}$ \\
Learning rate schedule & Linear warm-up followed by cosine decay \\
\bottomrule
\end{tabular}
\end{table}

\section{Fine-Tuning BTCNet on CHAT}
\label{app:finetuning_CHAT}
\paragraph{LoRA Adaptation Details.} When fine-tuning BTCNet on the CHAT dataset, LoRA adapters are applied to (i) the unimodal encoders corresponding to the selected modality pair and (ii) the Stage-2 cross-modal fusion transformer.

For the Stage-2 fusion model, LoRA is inserted into the attention projections and feed-forward layers (query/key/value projections, output projection, and MLP layers), using rank $r=64$, scaling factor $\alpha=128$, and dropout $0.05$.

For the unimodal encoders, LoRA is applied to the self-attention projections and MLP layers with the same configuration ($r=64$, $\alpha=128$, dropout $0.05$). 

\paragraph{Training Setup.}
The model architecture and loss formulation follow the Stage-2 configuration described in Table~\ref{tab:crossmodal_arch}. 
For CHAT fine-tuning, we adapt the optimization setup to reflect the smaller dataset size and the use of parameter-efficient LoRA adaptation.
Specifically, training is performed using an epoch-based schedule rather than a fixed iteration budget, corresponding to approximately 50 full passes over the CHAT training set.
We additionally use a larger effective batch size (128) and a higher learning rate ($3 \times 10^{-4}$), which are more suitable for LoRA-only updates, while keeping all non-LoRA parameters frozen.
These changes are relative to the Stage-2 pretraining setup summarized in Table~\ref{tab:crossmodal_train}.

\begin{table*}[tb]
\centering
\caption{
Comparison of the non--time-aware baseline BCNet and BTCNet after fine-tuning both models on the CHAT dataset.
Results are reported over three seeds.
Binary tasks report Accuracy (Acc), AUROC, and F1.
Sleep scoring reports Accuracy, weighted AUROC, and weighted F1. Results are averaged over three seeds and reported as mean (SD).
}
\label{tab:chat_finetune_time_vs_notime}
\small
\setlength{\tabcolsep}{6pt}

\begin{tabular}{l c l ccc ccc}
\toprule
Task & \%Pos & Modality Pair &
\multicolumn{3}{c}{BCNet (CHAT-FT)} &
\multicolumn{3}{c}{BTCNet (CHAT-FT)} \\
\cmidrule(r){4-6} \cmidrule(r){7-9}
& & &
\shortstack{Acc\\(\%)} &
\shortstack{AUROC\\(\%)} &
\shortstack{F1\\(\%)} &
\shortstack{Acc\\(\%)} &
\shortstack{AUROC\\(\%)} &
\shortstack{F1\\(\%)} \\
\midrule

Oxygen Desaturation
& 16.70
& EEG C4-M1 -- SPO2
& \shortstack{47.75\\{\scriptsize (0.37)}}
& \shortstack{59.69\\{\scriptsize (0.18)}}
& \shortstack{30.84\\{\scriptsize (0.14)}}
& \textbf{\shortstack{74.56\\{\scriptsize (1.73)}}}
& \textbf{\shortstack{74.41\\{\scriptsize (0.12)}}}
& \textbf{\shortstack{42.23\\{\scriptsize (0.24)}}} \\

\midrule

Hypopnea
& 3.78
& EEG C4-M1 -- SPO2
& \shortstack{78.47\\{\scriptsize (7.18)}}
& \shortstack{60.61\\{\scriptsize (1.05)}}
& \shortstack{9.83\\{\scriptsize (0.31)}}
& \textbf{\shortstack{91.88\\{\scriptsize (0.32)}}}
& \textbf{\shortstack{75.02\\{\scriptsize (0.05)}}}
& \textbf{\shortstack{20.24\\{\scriptsize (0.55)}}} \\

\midrule

Apnea--Hypopnea
& 5.92
& EEG C4-M1 -- SPO2
& \shortstack{66.90\\{\scriptsize (3.90)}}
& \shortstack{59.04\\{\scriptsize (0.73)}}
& \shortstack{13.64\\{\scriptsize (0.50)}}
& \textbf{\shortstack{88.72\\{\scriptsize (0.84)}}}
& \textbf{\shortstack{73.40\\{\scriptsize (0.16)}}}
& \textbf{\shortstack{24.87\\{\scriptsize (0.61)}}} \\

\midrule

Apnea
& 2.40
& EEG C4-M1 -- SPO2
& \shortstack{67.22\\{\scriptsize (3.27)}}
& \shortstack{56.35\\{\scriptsize (0.87)}}
& \shortstack{5.56\\{\scriptsize (0.26)}}
& \textbf{\shortstack{92.62\\{\scriptsize (1.39)}}}
& \textbf{\shortstack{70.01\\{\scriptsize (0.26)}}}
& \textbf{\shortstack{12.00\\{\scriptsize (0.11)}}} \\

\midrule

EEG-Arousal
& 8.07
& EEG O2-M1 -- EEG C4-M1
& \shortstack{69.97\\{\scriptsize (1.05)}}
& \shortstack{58.82\\{\scriptsize (0.25)}}
& \shortstack{17.56\\{\scriptsize (0.24)}}
& \textbf{\shortstack{88.76\\{\scriptsize (0.51)}}}
& \textbf{\shortstack{73.66\\{\scriptsize (0.12)}}}
& \textbf{\shortstack{33.01\\{\scriptsize (0.21)}}} \\

\midrule

5-stage Sleep Scoring
& --
& EEG C4-M1 -- SPO2
& \shortstack{31.30\\{\scriptsize (2.05)}}
& \shortstack{62.05\\{\scriptsize (0.09)}}
& \shortstack{29.88\\{\scriptsize (3.23)}}
& \textbf{\shortstack{49.53\\{\scriptsize (0.39)}}}
& \textbf{\shortstack{79.93\\{\scriptsize (0.05)}}}
& \textbf{\shortstack{49.03\\{\scriptsize (0.83)}}} \\

\bottomrule
\end{tabular}
\end{table*}

\section{Experiments and Results}
\subsection{Modality Screening Procedure}
\label{app:modality_screening}
Before evaluating BTCNet on downstream tasks, we identify
informative modality combinations using a fast screening
procedure. For each task on either NCH or CHAT, we randomly
sample a subset of examples (128{,}000 for NCH and 64{,}000
for CHAT) and extract frozen representations from the
pretrained encoders for all 120 possible modality pairs. We
then fit lightweight logistic regression classifiers using
scikit-learn~\cite{pedregosa2011scikit} with $\ell_2$
regularization and a small number of optimization iterations.
For binary tasks, we compute AUROC scores from predicted
class probabilities, while for multiclass tasks we report
weighted one-vs-rest AUROC. Modality pairs are ranked
according to these scores, and the top-performing
combinations are selected for subsequent evaluation. For binary classification tasks, we prioritize F1 binary score when ranking modality pairs, as it best reflects performance on rare clinical events. Under severe class imbalance, absolute F1 values are expected to be low and comparable to class prevalence for random predictions; thus, improvements in F1 indicate meaningful separability. AUROC is reported to show threshold-independent separability. This
procedure is used solely for relative ranking of modality
pairs and not for reporting final downstream performance. This procedure enables efficient screening of all 120
modality combinations for each task and dataset, facilitating
the identification of robust, task-specific modality pairs.

\subsection{Linear Probing Configurations}

Table \ref{app:tab:linear_probing} lists the configurations we used for fitting linear probes on the NCH and CHAT dataset.

\begin{table}[tb]
\centering
\caption{Linear probing configuration for downstream evaluation.}
\label{app:tab:linear_probing}
\small
\setlength{\tabcolsep}{6pt}
\begin{tabular}{l c}
\toprule
\textbf{Component} & \textbf{Setting} \\
\midrule
Probe architecture & Single linear layer \\
Loss function & Cross-entropy (class-weighted) \\
Optimizer & Adam \\
Batch size & 128 \\
Learning rate & $4 \times 10^{-3}$ \\
Weight decay & $1 \times 10^{-5}$ \\
Epochs (NCH) & 50 \\
Iterations per epoch (NCH) & 2000 \\
Epochs (CHAT) & 10 (full dataset pass) \\
\bottomrule
\end{tabular}
\end{table}

\subsection{Additional Comparisons: BTCNet vs BCNet on NCH}

\label{app:bcnet_vs_btcnet_additional_table}

Here, we report results from the remaining two out of three modalities that we experimented with to strengthen our claim that BTCNet outperforms its non-time counterpart BCNet on the NCH dataset. According to Table~\ref{tab:app_additional_modality_pairs}, across all evaluated tasks and modality pairs, BTCNet consistently improves AUROC and F1 score over the non–time-aware baseline, indicating better discrimination and robustness under class imbalance. While minor variations in accuracy are observed for a small number of channel combinations, accuracy is less informative in these highly imbalanced settings and does not reflect the improved minority-class detection captured by AUROC and F1 for binary tasks.

\begin{table*}[tb]
\centering
\caption{
Additional modality-pair results on the NCH dataset.
Binary tasks report Accuracy (Acc), AUROC, and F1.
Sleep scoring reports Accuracy, weighted AUROC, and weighted F1.
}
\label{tab:app_additional_modality_pairs}
\small
\setlength{\tabcolsep}{6pt}

\begin{tabular}{l c l ccc ccc}
\toprule
Task & \%Pos & Modality Pair &
\multicolumn{3}{c}{BCNet} &
\multicolumn{3}{c}{BTCNet} \\
\cmidrule(r){4-6} \cmidrule(r){7-9}
& & &
\shortstack{Acc\\(\%)} &
\shortstack{AUROC\\(\%)} &
\shortstack{F1\\(\%)} &
\shortstack{Acc\\(\%)} &
\shortstack{AUROC\\(\%)} &
\shortstack{F1\\(\%)} \\
\midrule

\multirow{2}{*}{Oxygen Desaturation}
& 8.78
& SPO2 -- RESP ABDOMINAL
& 88.48 & 78.83 & 38.78
& \textbf{90.16} & \textbf{81.59} & \textbf{44.58} \\
& 
& SPO2 -- EOG LOC-M2
& 88.05 & 78.79 & 38.48
& \textbf{89.83} & \textbf{81.93} & \textbf{44.11} \\

\midrule

\multirow{2}{*}{Hypopnea}
& 1.97
& EEG CZ-O1 -- SPO2
& \textbf{96.41} & 82.82 & 25.25
& 96.26 & \textbf{85.97} & \textbf{29.58} \\
& 
& EEG F3-M2 -- SPO2
& \textbf{96.64} & 82.83 & 25.50
& 96.33 & \textbf{86.30} & \textbf{30.52} \\

\midrule

\multirow{2}{*}{Apnea--Hypopnea}
& 2.80
& CAPNO -- SPO2
& 95.88 & 84.38 & 31.71
& \textbf{95.95} & \textbf{87.33} & \textbf{35.68} \\
& 
& EEG CZ-O1 -- SPO2
& \textbf{95.98} & 82.60 & 31.75
& 95.92 & \textbf{85.71} & \textbf{36.19} \\

\midrule

\multirow{2}{*}{Apnea}
& 0.83
& EEG O1-M2 -- SPO2
& 97.93 & 81.61 & 18.65
& \textbf{98.37} & \textbf{84.55} & \textbf{22.29} \\
& 
& EEG O2-M1 -- SPO2
& 97.78 & 81.62 & 18.41
& \textbf{98.24} & \textbf{84.53} & \textbf{21.53} \\

\midrule

\multirow{2}{*}{EEG-Arousal}
& 4.71
& RESP THORACIC -- EMG CHIN1--CHIN2
& 89.97 & 76.68 & 25.88
& \textbf{90.69} & \textbf{79.44} & \textbf{28.65} \\
& 
& EOG ROC-M1 -- EMG CHIN1--CHIN2
& 90.18 & 78.04 & 27.99
& \textbf{90.54} & \textbf{80.61} & \textbf{28.59} \\

\midrule

\multirow{2}{*}{5-stage Sleep Scoring}
& --
& EEG C3-M2 -- EOG LOC-M2
& 56.53 & 79.00 & 53.29
& \textbf{61.12} & \textbf{83.84} & \textbf{60.80} \\
& 
& EEG F4-M1 -- EOG ROC-M1
& 55.66 & 78.17 & 52.46
& \textbf{61.47} & \textbf{83.61} & \textbf{60.37} \\

\bottomrule
\end{tabular}
\end{table*}

\subsection{Linear Probing using SleepFM}

We follow the embedding extraction procedure described in~\cite{thapa2026multimodal} and keep all SleepFM encoder weights frozen throughout evaluation. Specifically, we map our original PSG channels into the modality groups expected by SleepFM and extract fixed-length embeddings from the pretrained model using 5-second windows.

Consistent with SleepFM’s modality definitions, we organize the available channels into three groups: \emph{BAS} (brain activity signals), \emph{RESP} (respiratory and oxygen-related signals), and \emph{EMG}. The BAS group consists of six EEG channels (C3–M2, O1–M2, O2–M1, C4–M1, F4–M1, F3–M2). The RESP group includes five respiratory-related signals (end-tidal CO$_2$, oxygen saturation, abdominal effort, chest effort, and snoring). The EMG group contains the chin electromyography channel. Channels without a clear functional correspondence to SleepFM’s predefined modalities are excluded from evaluation. Although SleepFM additionally supports ECG inputs, these signals are not available in either CHAT or NCH and are therefore omitted.

For each modality group, SleepFM produces a modality-specific temporal embedding sequence of shape $(T \times 128)$, where $T$ denotes the number of 5-second segments. Following prior work, we either aggregate embeddings by average pooling along the temporal dimension or flatten the temporal features directly to obtain a fixed-length representation for downstream linear probing, as described in the main paper

\begin{figure}[tb]
    \centering
    \includegraphics[width=0.9\linewidth]{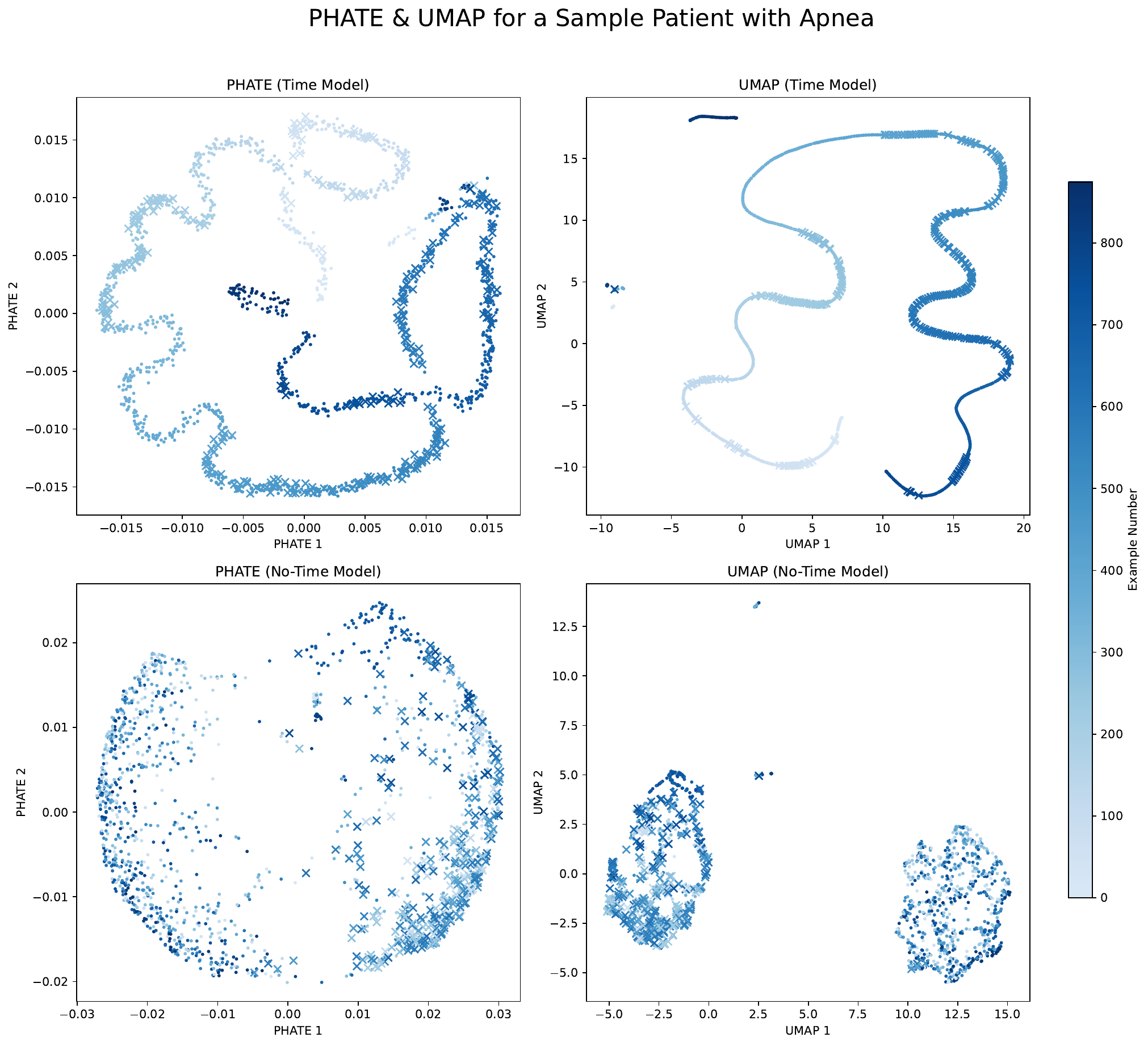}
    \caption{
    PHATE and UMAP embeddings for a representative patient with apnea using CAPNO and SPO2 channels.
    Positive events are indicated by $\times$ markers, while color denotes
    temporal order within the patient record.
    }
    \label{fig:phate-umap_1}
\end{figure}

\section{Additional PHATE Visualizations}
\begin{figure}[t]
    \centering
    \includegraphics[width=0.9\linewidth]{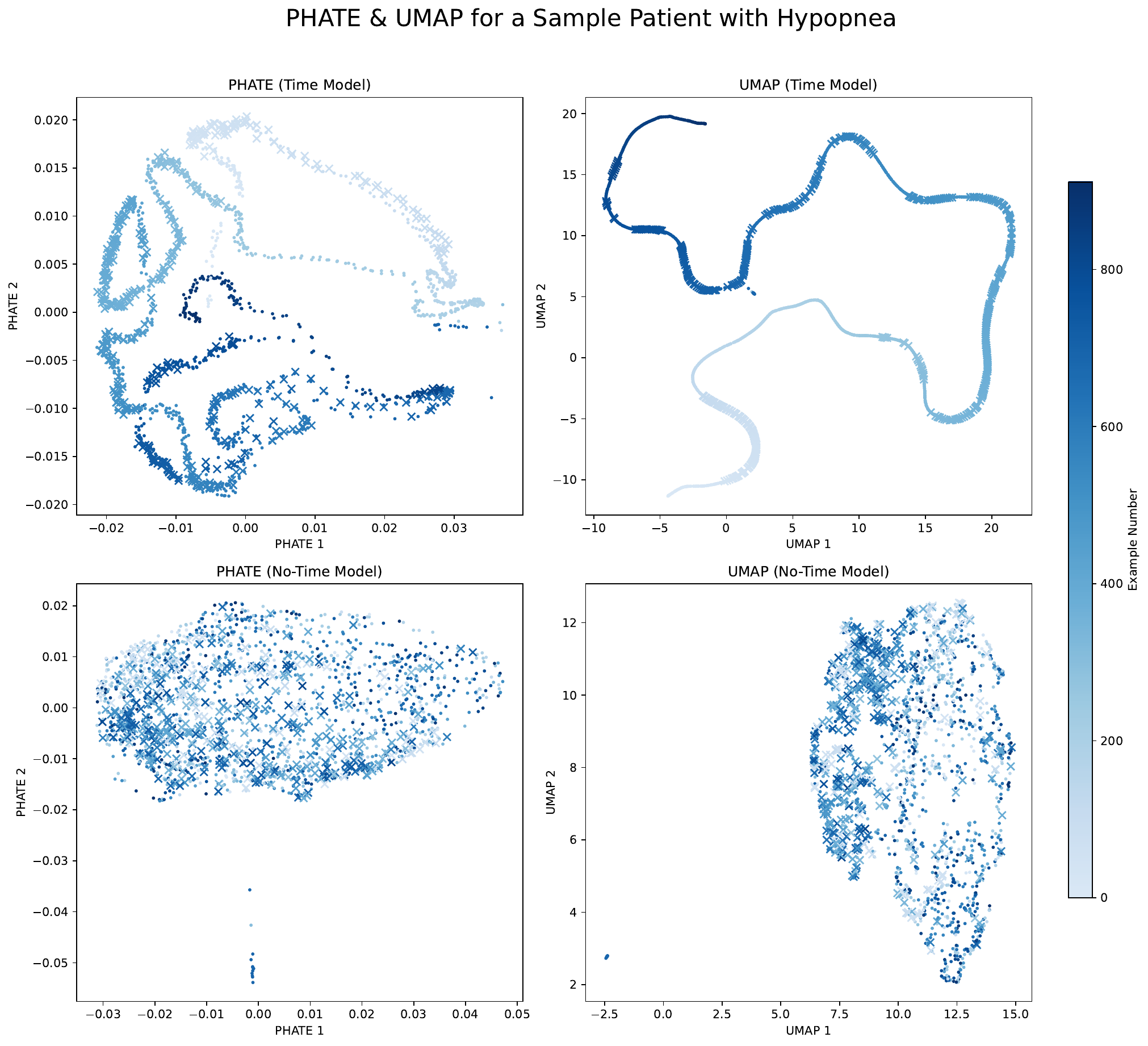}
    \caption{
    PHATE and UMAP embeddings for a representative patient with hypopnea using EEG O1-M2 and SPO2 channels.
    Positive events are indicated by $\times$ markers, while color denotes
    temporal order within the patient record.
    }
    \label{fig:phate-umap_2}
\end{figure}

To gain qualitative insight into how time-aware pretraining shapes the learned representation space, we visualize frozen embeddings from the BTCNet (time-aware) and BCNet (non-time-aware) models using PHATE, a nonlinear dimensionality reduction method that preserves both local neighborhood structure and global progression in high-dimensional data ~\cite{moon2019visualizing}. These visualizations are intended to provide intuition rather than quantitative evidence. Figure \ref{fig:phate} shows a representative example from a pediatric patient with frequent oxygen desaturation events, using the top-performing channel pair SPO2--CAPNO. The time-aware model exhibits a smoother and more coherent global structure, whereas the non-time-aware model produces more fragmented and scattered embeddings. This qualitative difference is consistent with our linear-probing results and provides intuition for why incorporating global time information leads to representations that are more amenable to separation in downstream tasks. Additional example PHATE plots, along with UMAP \cite{umap}, are shown in Figure \ref{fig:phate-umap_1} and Figure \ref{fig:phate-umap_2}.

\section{Expanded Background and Related Work}

\paragraph{Pediatric Sleep Analysis and Clinical Challenges.}

Despite increasing interest in automated PSG analysis, most computational sleep modeling efforts have focused on adult populations, while pediatric sleep remains comparatively understudied. This gap is partly due to the scarcity of publicly available pediatric datasets, limited software support for pediatric sleep event detection, and the underdiagnosis and underreporting of sleep disorders in children \cite{sleep_textbook, blunden2004sleep}. Importantly, pediatric sleep physiology differs substantially from adult sleep in terms of sleep architecture, respiratory control, and autonomic regulation \cite{accardo2010differences, owens2012pro}. As a result, models trained on adult sleep data often generalize poorly to pediatric populations, even for foundational tasks such as sleep stage classification \cite{nazih2023influence}.

\paragraph{Supervised and Self-Supervised Learning for Sleep Analysis.}
To reduce reliance on expensive expert annotations, a wide range of machine learning approaches have been explored for automated sleep assessment. However, the majority of existing methods rely on supervised learning \cite{UNet, wenjian2023dynamicsleepnet, eldele2021attention, lee2022automatic}, which requires large volumes of labeled data that are difficult to obtain in clinical practice. Self-supervised learning (SSL) offers a promising alternative by learning representations from unlabeled PSG recordings using pretext objectives such as masked signal reconstruction, which can then be transferred to downstream tasks at substantially lower annotation cost.

While SSL has been widely adopted in other domains, its application to pediatric sleep modeling remains limited. Moreover, most existing SSL-based approaches in sleep analysis are designed primarily for sleep stage classification \cite{UNet, wenjian2023dynamicsleepnet, lee2024neuronet, kostas2021bendr}. Consequently, SSL methods targeting clinically important diagnostic tasks, such as apnea, hypopnea, and oxygen desaturation detection, remain relatively underexplored, despite their strong associations with cardiovascular and neurological outcomes \cite{yaggi2010adult}.

\paragraph{Self-Supervised Learning for EEG and Multimodal Sleep Data.}
Influential SSL approaches include masked modeling methods such as MAE \cite{he2022masked} in vision and BERT \cite{devlin2019bert} in language, as well as contrastive learning frameworks such as SimCLR \cite{chen2020simple} and MoCo \cite{he2020momentum}. Multimodal extensions, including CLIP \cite{radford2021learning}, demonstrate the effectiveness of contrastive alignment across heterogeneous modalities.

In the sleep domain, SSL efforts have predominantly focused on uni-modal EEG-based representation learning, with an emphasis on sleep stage classification. BENDR \cite{kostas2021bendr} adapts masked self-supervised objectives to raw EEG signals, while EEGPT \cite{wang2024eegpt} introduces spatio-temporal alignment and hierarchical modeling to improve representation quality. NeuroNet \cite{lee2024neuronet} further combines masked prediction and contrastive learning on single-channel EEG with temporal modeling across epochs.

More recent work has explored multimodal SSL for sleep analysis. COCOA \cite{deldari2022cocoa} introduces cross-modality contrastive learning for multimodal physiological time series, while SleepFM \cite{sleepfm_original} proposes a multimodal foundation model using leave-one-out contrastive alignment across modalities. SynthSleepNet \cite{CASleepNet} extends hybrid SSL to multimodal PSG data by integrating masked prediction and contrastive learning with temporal context modeling. However, multimodal SSL techniques trained specifically on pediatric sleep data remain scarce. To our knowledge, PedSleepMAE \cite{pandey2024pedsleepmae} is the only existing SSL framework trained on large-scale pediatric PSG data (NCH Sleep Databank) for sleep stage classification and related diagnostic tasks.

\end{document}